\documentclass[12pt]{article}

\usepackage{amsmath,amssymb,amsfonts}
\usepackage{bbm}
\usepackage[justification=centering,margin=10pt,font=small,labelfont=bf]{caption}
\usepackage{graphicx,xcolor}

\usepackage{apacite}
\usepackage[comma,authoryear,round]{natbib}
\usepackage{hyperref}
\begin{document}

\noindent \textbf{Data-driven advice for interpreting local and global model predictions in bioinformatics problems}
\vskip 7mm

\noindent Markus Loecher

\noindent
Berlin School of Economics and Law,\\
10825 Berlin, Germany\\
E-mail: markus.loecher@hwr-berlin.de\\

\vskip 5mm

\noindent Wu Qi

\noindent
Dept. of Statistics\\
Humboldt University\\
E-mail: wuqiqi0704@gmail.com

\begin{abstract}
Tree-based algorithms such as random forests and gradient boosted trees continue to be among the most popular and powerful machine learning models used across multiple disciplines.
The conventional wisdom of estimating the impact of a feature in tree based models is to measure the \textit{node-wise reduction of a loss function}, which (i) yields only global importance measures and (ii) is known to suffer from severe biases. Conditional feature contributions (CFCs) \cite{saabas2019treeInterpreter} provide \textit{local}, case-by-case explanations of a prediction by following the decision path and attributing changes in the expected output of the model to each feature along the path.
However, \citep{lundberg2020local} pointed out a potential bias of CFCs which depends on the distance from the root of a tree. The by now immensely popular alternative, \textit{SHapley Additive exPlanation} (SHAP) values appear to mitigate this bias but are computationally much more expensive. \\
Here we contribute a thorough comparison of the explanations computed by both methods on a set of 164 publicly available classification problems in order to provide data-driven algorithm recommendations to current researchers. 
For random forests, we find extremely high similarities and correlations of both local and global SHAP values and CFC scores, leading to very similar rankings and interpretations. 
Analogous conclusions hold for the fidelity of using global feature importance scores as a proxy for the predictive power associated with each feature.
\end{abstract}

\noindent Key Words: variable importance, random forests, trees, Gini impurity


\section{Interpreting Model Predictions}

Tree-based algorithms such as random forests and gradient boosted trees continue to be among the most popular and powerful machine learning models used across multiple disciplines.

While variable importance is not easily defined as a concept ~\citep{gromping2009variable}, the conventional wisdom of estimating the impact of a feature in tree based models is to measure the \textit{node-wise reduction of a loss function}, which (i) yields only global importance measures and (ii) is known to suffer from severe biases.
Nevertheless, variable importance measures for random forests have been receiving increased attention in bioinformatics, for instance to select a subset of genetic markers relevant for the prediction of a certain disease.
They also have been used as screening tools \citep{diaz2006gene,menze2009comparison}
in important applications, highlighting the need for reliable and well-understood feature importance
measures.

The default choice in most software implementations \citep{randomForest2002} of random forests \citep{Breiman2001} is the \textit{mean decrease in impurity (MDI)}. The MDI of a feature is computed as a (weighted) mean of the individual trees' improvement
in the splitting criterion produced by each variable. A substantial shortcoming of this default measure is its evaluation on the in-bag samples, which can lead to severe overfitting and bias \citep{kim2001classification,Strobl2007a,loecher2020unbiased,loecher2020unbiasedArxiv}. \\

\subsection{Conditional feature contributions (CFCs) \label{sec:TreeInterpreter}}

The conventional wisdom of estimating the impact of a feature in tree based models is to measure the \textbf{node-wise reduction of a loss function}, such as the variance of the output $Y$, and compute a weighted average of all nodes over all trees for that feature. By its definition, such a \textit{mean decrease in impurity} (MDI) serves only as a global measure and is typically not used to explain a \textit{per-observation, local impact}.
~\cite{saabas2019treeInterpreter} proposed the novel idea of explaining a prediction by following the
decision path and attributing changes in the expected output of the model to each feature along the path.

Let $f$ be a decision tree model, $x$ the instance we are going to explain, $f(x)$ the output of the model for the current instance, and $f_x(S) \approx E[f(x) \mid x_S]$ the estimated expectation of the model output conditioned on the set $S$ of feature values, then -following ~\cite{lundberg2019consistent}- we can define the {\it Saabas value}\footnote{synonymous with {\it conditional feature contributions}} for the $i$'th feature as

\begin{equation}
\phi^{s}_i(f,x) = \sum_{j \in D^i_x} f_x(A_j \cup j) - f_x(A_j), 
\label{eq:saabas}
\end{equation}

\noindent where $D^i_x$ is the set of nodes on the decision path from $x$ that split on feature $i$, and $A_j$ is the set of all features split on by ancestors of $j$. Equation~\ref{eq:saabas} results in a set of feature attribution values that sum up to the difference between the expected output of the model and the output for the current prediction being explained. When explaining an ensemble model made up of a sum of many decision trees, the CFCs for the ensemble model are defined as the sum of the CFCs for each tree.

 
In the light of wanting to explain the predictions from tree based machine learning models,  these \textit{conditional feature contributions}  are rather appealing, because
\begin{enumerate}
  \item The positive and negative contributions from nodes convey directional information unlike the strictly positive purity gains.
  \item By combining many local explanations, we can represent global structure while retaining local faithfulness to the original model.
  \item The expected value of every node in the tree can be estimated efficiently by averaging the model output over all the training samples that pass through that node.
  \item The algorithm has been implemented and is easily accessible in a python \citep{saabas2019treeInterpreter} and R \citep{tree.interpreter2020} library.
\end{enumerate}

\subsection{SHAP values \label{sec:TreeExplainer}}

However, ~\cite{lundberg2020local} pointed out  that it is strongly biased to alter the impact of features based on their distance from the root of a tree. This causes CFC scores to be inconsistent, which means one can modify a model to make a feature clearly more important, and yet the CFC attributed to that feature will decrease.
As a solution, the authors developed an algorithm (``TreeExplainer'') that computes local explanations based on exact Shapley values in polynomial time\footnote{A python library is available at \url{https://github.com/slundberg/shap}.}. 
As explained in  ~\cite{lundberg2019consistent}, Shapley values are computed
by introducing each feature, one at time, into a conditional expectation function of the model's output, $f_x(S) \approx E[f(x) \mid x_S]$, and attributing the change produced at each step to the feature that was introduced; then averaging this process over all possible feature orderings. Shapley values represent the only possible method in the broad class of {\it additive feature attribution methods} that will simultaneously satisfy three important properties: {\it local accuracy}, {\it consistency}, and {\it missingness}:
\begin{equation}
\phi_i(f,x) = \sum_{R \in \mathcal{R}} \frac{1}{M!} \left [ f_x(P^R_i \cup i) - f_x(P^R_i) \right ]
\label{eq:shapley}
\end{equation}
where $\mathcal{R}$ is the set of all feature orderings, $P^R_i$ is the set of all features that come before feature $i$ in ordering $R$, and $M$ is the number of input features for the model.

One should not forget though that the same idea of adding \textit{conditional feature contributions} lies at the heart of  \textit{TreeExplainer} with one important difference.
While SHAP values average the importance of introducing a feature over all possible feature orderings, CFC scores only consider the single ordering defined by a tree's decision path. 
~\cite{lundberg2019consistent} warns. "But since they do not average over all orderings, they do not match the SHAP values, and so must violate consistency."

\section{SHAP versus CFCs}

The main contribution of this paper is a thorough, direct, empirical comparison of CFC and SHAP scores for two tree ensemble methods: random forests (RF) and gradient boosted trees (XGboost). (All our simulations utilize the \textit{sklearn} library, in particular the methods \textit{RandomForestClassifier} and \textit{XGBClassifier}.)
Our initial focus in section \ref{sec:localComp} are the local explanations, while section \ref{sec:predPow} provides a different perspective on the global predictive discernability. 
We are first going to describe the data sets used in our study.

\subsection{Penn Machine Learning Benchmark (PMLB) \label{sec:PMLB}}

The algorithms were compared on 164 supervised classification datasets from the Penn Machine Learning Benchmark (PMLB) \citep{olson2017data}. PMLB is a collection of publicly available classification problems that have been standardized to the same format and collected in a central location with easy access via Python\footnote{URL: \url{https://github.com/EpistasisLab/penn-ml-benchmarks}}.
~\cite{olson2017data} compared 13 popular ML algorithms from scikit-learn\footnote{The entire experimental design consisted of over 5.5 million ML algorithm and parameter evaluations in total.} and found that Random Forests and boosted trees ranked consistently at the top.

\subsection{Comparative Study: local explanations\label{sec:localComp}}
\begin{figure}[!htbp]
  \centering
  \includegraphics[width=0.50 \textwidth]{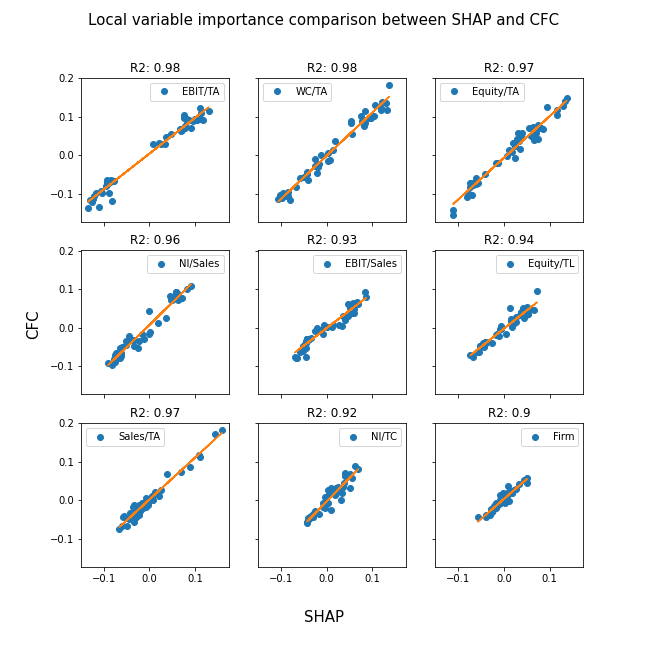}%
  \caption{Scatter plots of local SHAP and CFC values for dataset `analcatdata japansolvent', a linear regression is fitted to each variable and the corresponding R-Squared values are reported}
  \label{fig:local_shapcfc}
\end{figure}

Out of the 164 datasets in PMLB collection, one dataset `analcatdata japansolvent' is chosen here to illustrate the local variable importance comparison between SHAP and CFC. It is a binary classification task, the dataset consists of 9 numerical variables and 52 observations. A more detailed description about each dataset is provided by the authors\footnote{\url{https://epistasislab.github.io/pmlb/}}. The scatter plots shown in Figure \ref{fig:local_shapcfc} demonstrate the strong correlation between the variable importance calculated by SHAP and CFC methods for random forests. The plots are ranked by the (global) SHAP importance values, i.e. the first variable EBIT/TA on the top left graph is the most important variable for the classification task according to SHAP, and the second variable WC/TA  is the second important, and so forth. Additionally, a linear regression is fitted to each scatter plot, the corresponding R-Squared value of each linear regression is shown above each scatter plot for each variable, which confirms the strong linear correlation between SHAP and CFC computed variable importance. 

\begin{figure}[!htbp]
  \centering
  \includegraphics[width=0.45 \textwidth]{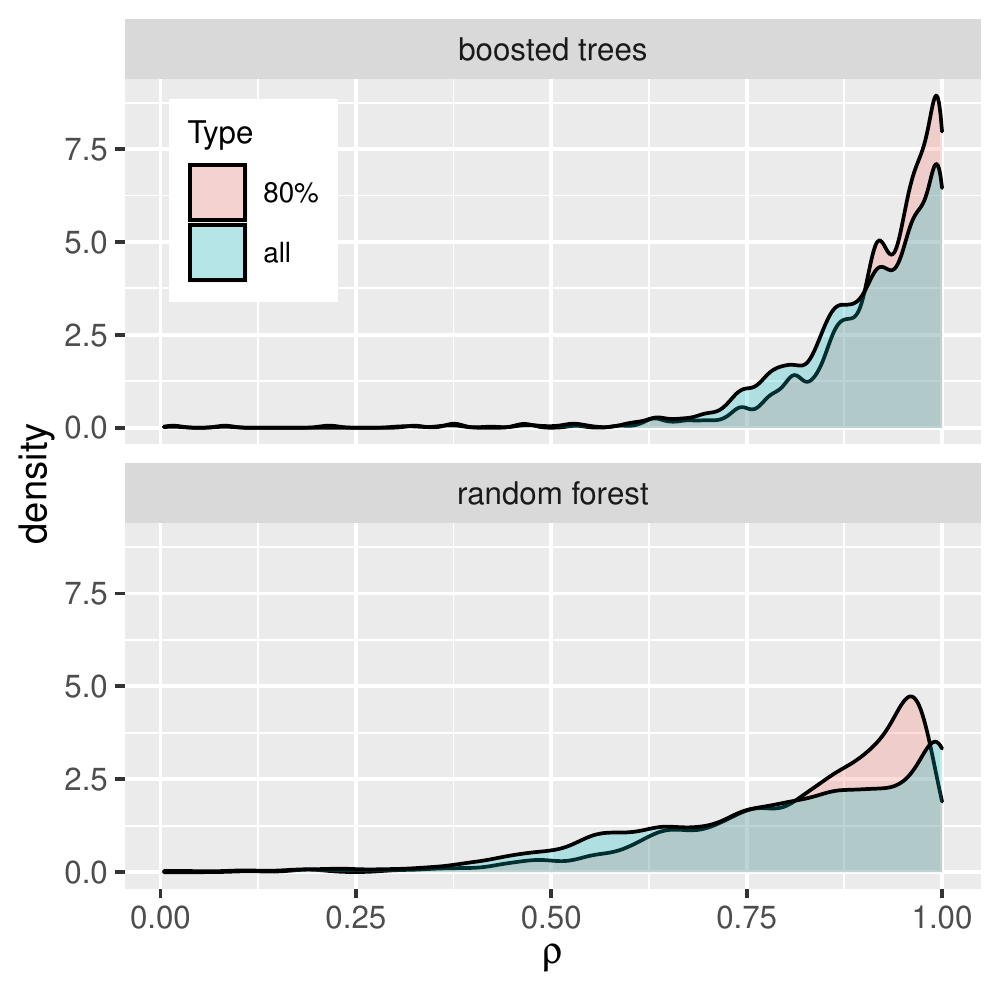}%
  \caption{Distribution of correlation coefficients between CFC and SHAP scores, one for each variable in each dataset. Filtered(80\%) stands for variables that contribute to $80\%$ of the total importance for each dataset. }
  \label{fig:global_shapcfc}
\end{figure}

In addition to this anecdotal example of the local variable importance comparison between SHAP and CFC, Figure \ref{fig:global_shapcfc} depicts the general distribution of all local correlation between SHAP and CFC scores in all PMLB datasets. For both RF and XGboost very strong correlations between SHAP and CFC are observed. The density plots display all correlations as well as a filtered distribution, including only variables that contribute $80\%$ to the total importance for each dataset. We consider this filtering criterion meaningful, as the correlation calculation of SHAP and CFC might not be highly representative for variables with a low importance score. We mention in passing that about half of the variables are filtered out. We also observe that there's a peak correlation value $1$ in the total correlation distribution plot, which is due to the bias of low/zero importance variables.  

\subsection{Predictive Power of Feature Subsets\label{sec:predPow}}

In the previous section, we compared the local explanations provided by two rather different algorithms. Since these data sets are not generated by simulations, we do not really have any ground truth on the "importance" or relevance of each feature's contribution to the outcome. In fact, ~\cite{loecher2020unbiasedArxiv} showed that even SHAP and CFC scores suffer from the same bias observed in the original MDI. Hence, we cannot really quantify which score provides more meaningful or less misleading measures.
At least for the global scores, we can inquire whether the sum of a method’s importance scores is a reliable proxy for the predictive power of subsets of features.
\begin{figure}[!htbp]
  \centering
  \includegraphics[width=0.47 \textwidth]{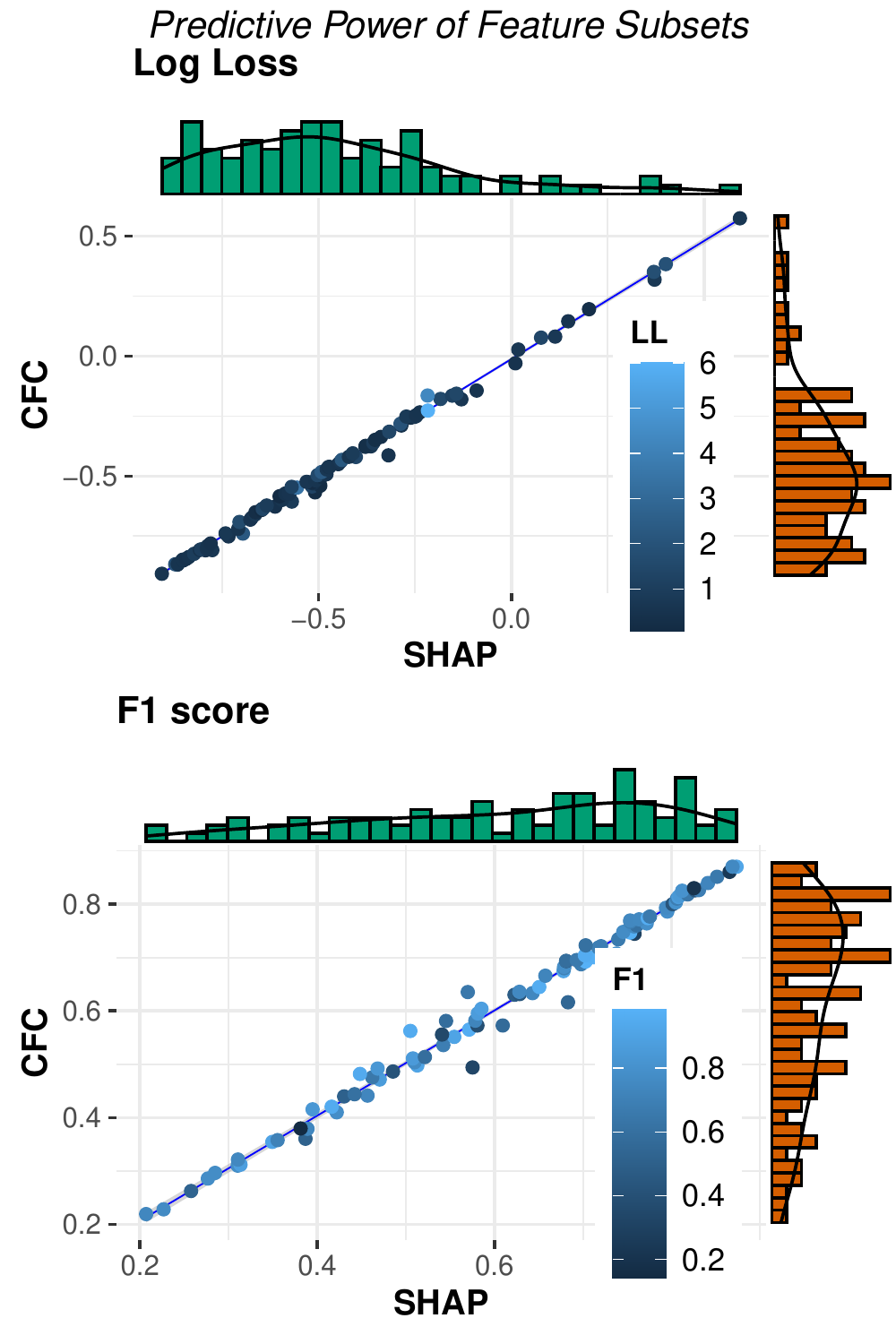}%
  \caption{For both CFC and SHAP we measured the correlation between the model loss and the total importance of the included features. Here, loss is measured either as cross entropy ("log loss", left panel) or as 1 minus the F1 score (right panel).  }
  \label{fig:PredictiveLoss_LL_F1}
\end{figure}
We chose to replicate the experiments in ~\cite{NEURIPS2020_covert}: "We generated several thousand feature subsets for each dataset, re-trained models for each subset, and then measured the correlation between the model loss and the total importance of the included features". In particular, the feature subsets were independently sampled as follows: sample the subset's cardinality $k \in {0, 1, \ldots, d}$ uniformly at random, then select $k$ elements from ${1, \ldots, d}$ uniformly at random. We computed both SHAP/CFC and the retrained models' loss (either log loss or the inverse F1 score) on a test set.
This strategy yields one correlation coefficient for SHAP and CFC per data set.
Figure~\ref{fig:PredictiveLoss_LL_F1} depicts these correlations for random forests as a scatterplot with marginal distributions \citep{ggstatsplot2021}.
Analogous results for boosted trees are shown in Figure~\ref{fig:PredictiveLoss_LL_F1_boosting} .
\begin{figure}[!htbp]
  \centering
  \includegraphics[width=0.47 \textwidth]{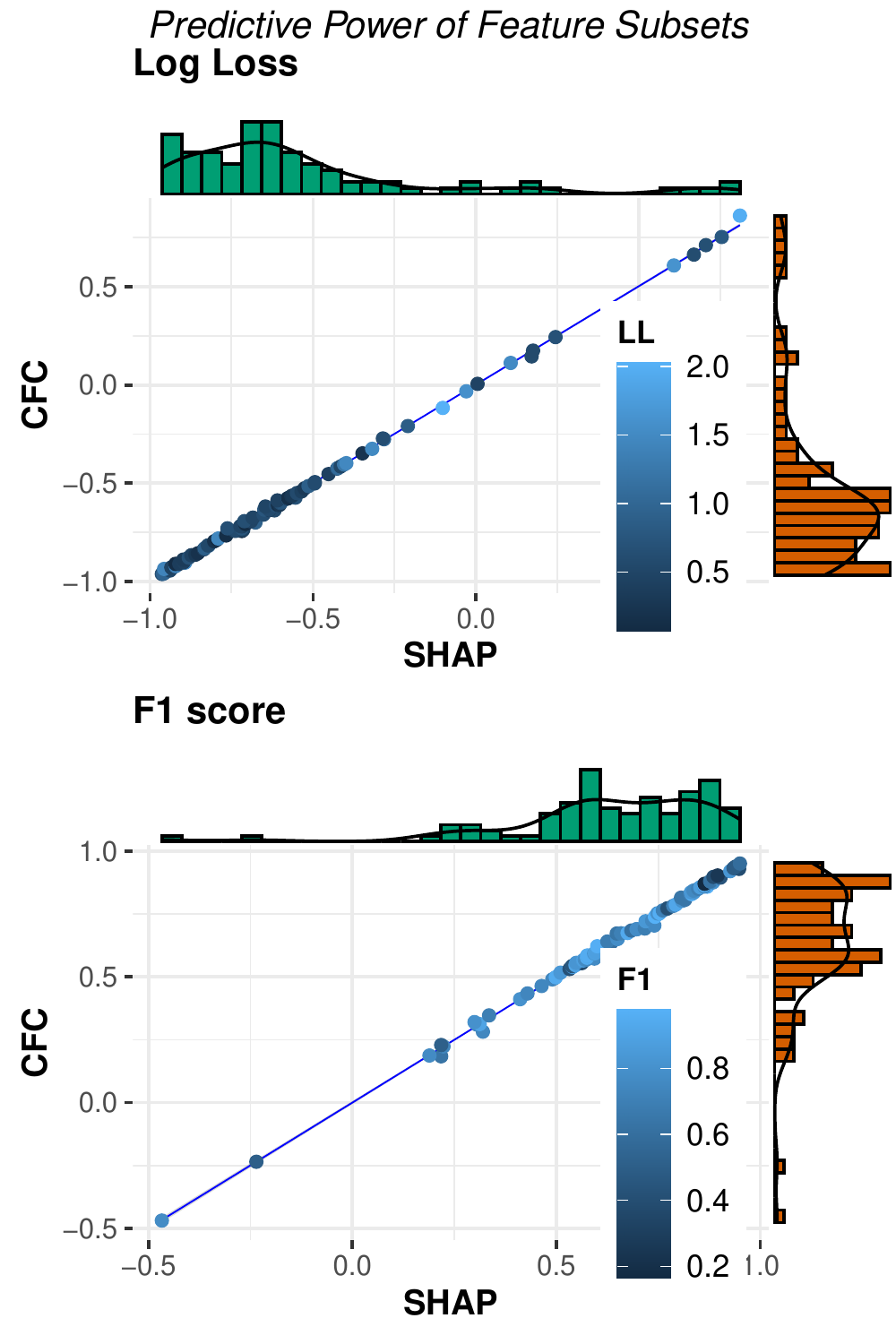}%
  \caption{ Same as Figure~\ref{fig:PredictiveLoss_LL_F1} but for boosted trees.}
  \label{fig:PredictiveLoss_LL_F1_boosting}
\end{figure}

For both choices of loss function and across all data sets, the measured correlations are almost identical for the global SHAP and CFC scores. The strength of this association surprised us and suggests that the global CFC scores measure the predictive information content of a features with as much fidelity as the global SHAP values.
We mention in passing the (albeit few) surprisingly positive correlations between the log loss and feature importances suggesting that sometimes including features with high importance scores can hurt model performance.
We speculate that this counter-intuitive phenomenon is due to the overfitting/bias of SHAP and MDI scores reported in ~\cite{loecher2020unbiasedArxiv}.

\section{Discussion}

The rapidly rising interest in explainable AI and interpretable machine learning has made it very clear that local explanations for complex models have a significant impact for most applications in general and particularly so for highly regulated domains, such as healthcare, finance and public services.
Explanations can provide transparency to customers and overseeing agencies, help find biases in algorithmic decision making, facilitate human–AI collaboration, and could even speed up model development and debugging.
The  increasingly popular SHAP scores play a vital role in this context since they fulfil certain optimality requirements borrowed from game theory and -for trees only- offer somewhat efficient computational shortcuts.
~\cite{lundberg2020local} convincingly showcases the value of local explanations in the realm of medicine.
SHAP values are used to (1) identify high-magnitude but low-frequency nonlinear mortality risk factors in the US population, (2) highlight distinct population subgroups with shared risk characteristics, (3) identify nonlinear interaction effects among risk factors for chronic kidney disease. Furthermore, "they enable doctors to make more informed decisions rather than blindly trust an algorithm’s output".

\noindent
The main contribution of this paper is to demonstrate empirically the effective similarity of SHAP scores and the much simpler conditional feature contributions which are defined for trees only.
Our findings are significant not only because CFCs are computationally cheaper by orders of magnitude but also because of their close connection to the conventional feature importance measure MDI as derived in ~\cite{loecher2020unbiased}. 
Why is this important? It could help explain why SHAP scores suffer from the same bias in the presence of predictors with greatly varying number of levels. 
For example, ~\cite{loecher2020unbiasedArxiv} demonstrated that both the CFCs as well as the SHAP values are highly susceptible to “overfitting” to the training data and proposed a correction based on oob data.

\noindent
For convenience we restricted our analysis to a large and diverse set of classification problems only but initial explorations indicate that the main conclusion holds for regression trees as well.
Going beyond random forests and boosting, we also found similar results for single classification trees.

\section{Acknowledgements}

We thank ~\cite{olson2017data} for providing the complete code\footnote{\url{https://github.com/rhiever/sklearn-benchmarks}} required both to conduct the algorithm and hyperparameter optimization study, as well as access to the analysis and results.
We also thank the authors of \cite{NEURIPS2020_covert} for providing us details on their simulations.


\end{document}